\documentclass[]{acmart}
\AtBeginDocument{%
  \providecommand\BibTeX{{%
    \normalfont B\kern-0.5em{\scshape i\kern-0.25em b}\kern-0.8em\TeX}}}

\setcopyright{acmcopyright}
\copyrightyear{2022}
\acmYear{2022}
\acmDOI{XXXXXXX.XXXXXXX}

\acmConference[Ubicomp22]{ UbiComp/ISWC '22: Adjunct Proceedings of the 2022 ACM International Joint Conference on Pervasive and Ubiquitous Computing and Proceedings of the 2022 ACM International Symposium on Wearable Computers}{Sep 11--15, 2022}
\begin{document}

\title{Autonomous Delivery of Multiple Packages Using Single Drone in Urban Airspace}

\author{Seunghyun Lee}
\affiliation{%
  \institution{The University of Sydney}
  \streetaddress{City Road Camperdown/Darlington}
  \city{Sydney}
  \country{Australia}}
\email{slee3812@uni.sydney.edu.au}

\author{Babar Shahzaad}
\affiliation{%
\institution{The University of Sydney}
  \streetaddress{City Road Camperdown/Darlington}
  \city{Sydney}
  \country{Australia}
}
\email{babar.shahzaad@sydney.edu.au}

\author{Balsam Alkouz}
\affiliation{%
  \institution{The University of Sydney}
  \streetaddress{City Road Camperdown/Darlington}
  \city{Sydney}
  \country{Australia}}
\email{balsam.alkouz@sydney.edu.au}

\author{Abdallah Lakhdari}
\affiliation{%
 \institution{The University of Sydney}
  \streetaddress{City Road Camperdown/Darlington}
  \city{Sydney}
 \country{Australia}}
\email{abdallah.lakhdari@sydney.edu.au}

\author{Athman Bouguettaya}
\affiliation{%
 \institution{The University of Sydney}
  \streetaddress{City Road Camperdown/Darlington}
  \city{Sydney}
 \country{Australia}}
\email{athman.bouguettaya@sydney.edu.au}

\renewcommand{\shortauthors}{Lee and S., et al.}

\begin{abstract}
  Current drone delivery solutions mainly focus on single package delivery using one drone. However, the recent developments in drone technology enable a drone to deliver multiple packages in a single trip. We use the nearest destination first strategy for the faster delivery of packages in a skyway network. This demonstration is a proof-of-concept prototype for the multi-package delivery in urban airspace following a skyway network. We deploy and test this multi-package drone delivery in an indoor testbed environment using a 3D model of Sydney CBD. Demo: \url{https://youtu.be/YTwsIfUvWPc}
\end{abstract}





\maketitle

\section{Introduction}

Drones are becoming increasingly important for a wide range of commercial applications in urban areas \cite{shakhatreh2019unmanned}. Examples of these applications include public security, remote sensing, surveillance, photography, and delivery of goods \cite{mohsan2022towards}. The continual growth of e-commerce, especially during the COVID-19 pandemic, has revolutionized the way customers acquire goods and services \cite{nanda2021would}. The ubiquity of drones in the sky has prompted an increasing interest of several e-commerce companies such as UPS, Flytrex, and Amazon Prime Air to use drones for package delivery \cite{aurambout2019last}. Several countries have used drones for safe and contactless deliveries during the pandemic lockdowns \cite{shahzaad2021robust}. Drone delivery is highly desired in urban areas to reduce delivery time and traffic congestion on roads by utilizing urban airspace \cite{alkouz2021service}.

The recent developments in drone technology show that drones can carry \textit{multiple packages} \cite{shahzaad2022service}. Therefore, a drone can serve more than one customer in one trip. For example, the Freefly Alta X\footnote{\url{https://freeflysystems.com/alta-x/specs}} drone has a maximum payload capacity of up to 15.9 kg. The majority of Amazon's delivery items (86\%) are less than 2.27 kg \cite{doi:10.1111/drev.10313}. In this respect, the drones from the company mentioned above can \textit{deliver multiple items} from Amazon in a single trip. The added benefits of using a single drone for multi-package delivery include reduced cost per delivery, reduced congestion in the sky, and reduced number of trips back to the depot.

Certain technological and regulatory challenges hinder the potential utilization of drones in urban airspace \cite{janszen2022constraint}. These challenges include the limited flying range, battery capacity, and flight regulations such as flying within line of sight and avoiding no-fly zones or restricted areas \cite{shahzaad2020game}. A skyway network enables the real-world deployment of drone delivery systems in urban airspace addressing the challenges mentioned above \cite{10.1145/3460418.3479289}. A {\em skyway network} is defined as a set of skyway \textit{segments} that directly connect two nodes representing take-off and landing stations \cite{shahzaad2021top}. Take-off and landing stations (aka network nodes) are typically from and to building rooftops.

\begin{figure}[t]
\centering
\includegraphics[width=0.5\linewidth]{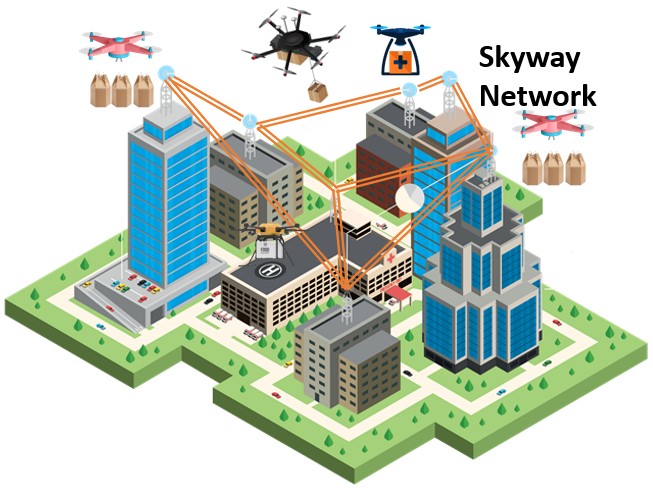}
   
\caption{Multi-Package Delivery in a Skyway Network}
\label{multi-package-skyway}

\end{figure}

This paper focuses on the design and proof-of-concept demonstration of multi-package delivery using a single drone in one trip in urban airspace. Fig. \ref{multi-package-skyway} depicts the delivery of multiple packages using a single drone in a skyway network. We use the building rooftops as pickup and delivery locations for drones. Several approaches exist that focus on optimizing the drone-based multi-package delivery by proposing payload-mass-aware trajectory planning \cite{tsoutsouras2020payloadmassaware} and load-dependent flight speed-aware \cite{9052143} drone delivery. In this demonstration, we use the Nearest Destination First (NDF) strategy, which delivers packages to the closest destination first.

\section{Demo Setup}

\subsection{Indoor Testbed Environment}

Having an outdoor test environment is fraught with risks because of government restrictions when using drones in urban areas \cite{jones2017international}. We use a Crazyflie 2.1 nano-quadcopter drone by Bitcraze\footnote{\url{https://www.bitcraze.io/products/crazyflie-2-1/}} in a 3D model of the Sydney CBD as an indoor testbed to mimic a skyway network (Fig. \ref{skyway}). We use a \emph{Crazyflie} drone because it is safe and well suited for the indoor testbed due to its small size and robust nature. One HTC Vive base station is used that employs infrared laser technology to calculate the coordinates of the drone. We compute a path using the NDF strategy to perform drone-based deliveries from a given source to respective destinations.

\subsection{Multi-Package Delivery Mechanism}

\begin{figure}
    \centering
    \begin{minipage}{0.5\textwidth}
        \centering
        \includegraphics[width=\linewidth]{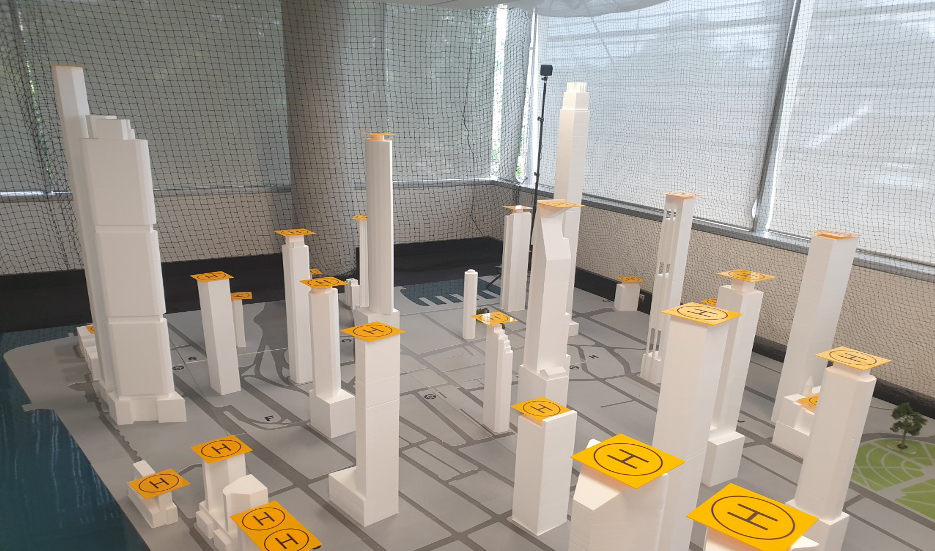} % first figure itself
        \caption{3D Model of Sydney CBD}
        \label{skyway}
    \end{minipage}\hfill
    \begin{minipage}{0.5\textwidth}
        \centering
        \includegraphics[width=0.7\linewidth, height=4.5cm]{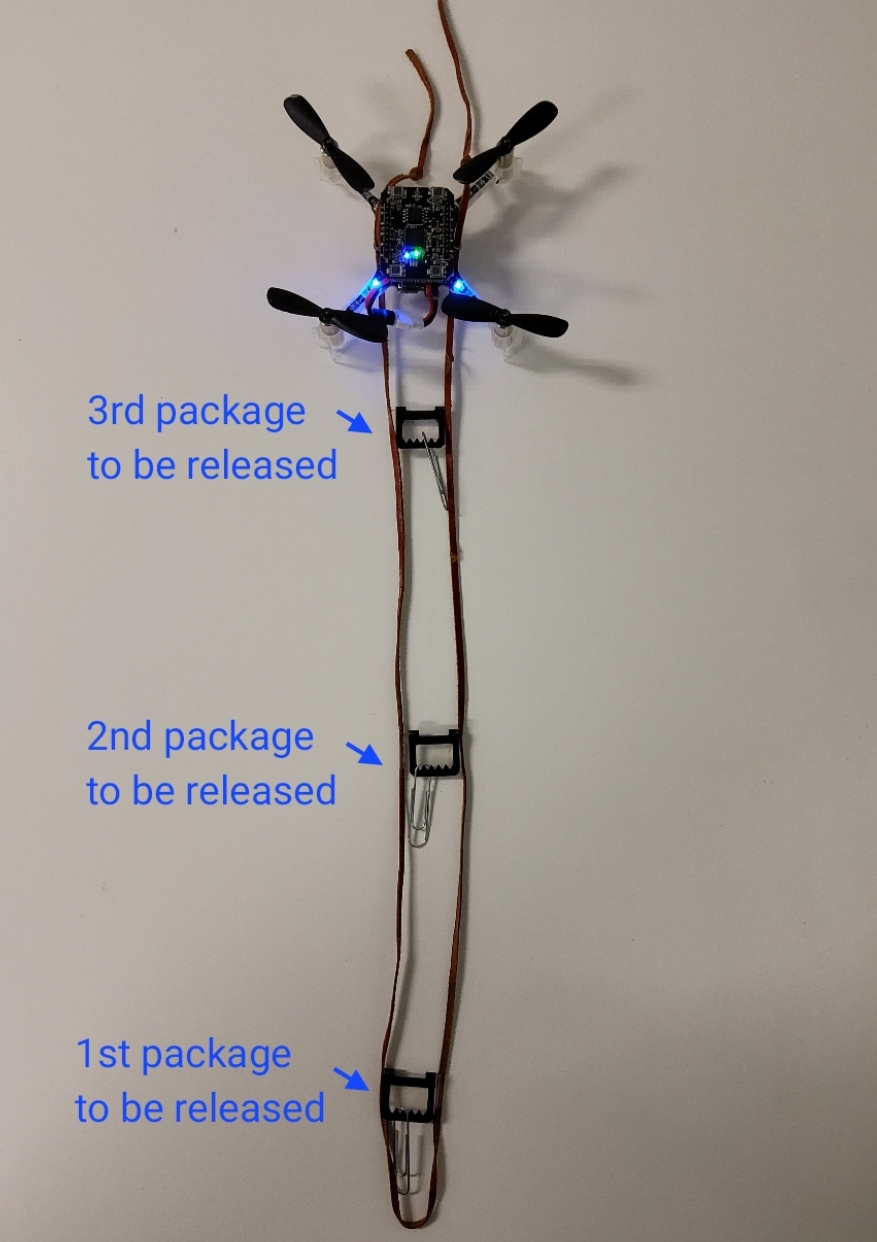} % first figure itself
        \caption{Multi-Level Hanging Mechanism}
        \label{string-mechanism}
    \end{minipage}
\end{figure}

We envision two different mechanisms for multi-package delivery using a single drone: (1) Using electromagnets and (2) Using multi-level hanging. The first mechanism uses electromagnets to lift and drop the packages at the same height. In this case, we can release the package by turning off the relevant electromagnet. However, the lightest electromagnet available on the market weighs more than the maximum payload capacity of the Crazyflie drone. Therefore, we use the multi-level hanging mechanism for delivering multiple packages. We use a string to hang packages at different levels. Each end of the string is tied to the middle of the drone's sides, forming a ring shape to maintain the drone's stability (Fig. \ref{string-mechanism}). We tape around the string at three different levels allowing the string and tape to form a ladder shape which serves as a hanger for the package. A hook-shaped frame hanger with a clip is used as a package that weighs within the maximum payload capacity of the drone.

\section{Autonomous Navigation for Multi-Package Drone Delivery}

We use the NDF strategy to navigate the drone autonomously for delivering multiple packages. The initial position of the drone is set to (0, 0, 0) coordinates in the 3D coordinate system. The drone starts the delivery operation carrying three packages. These packages are hung at drones on different levels by a string. The order of package levels is reverse to that of release, which depends on the distance from the start node to its destination node under NDF strategy. For example, the package hanging at the top level is dropped to the final delivery target. We fly the drone at a height above the delivery target's total height (i.e., building rooftop) and the length of the string. This ensures that no package hits the building rooftop and drops on an undesired location. When the drone reaches a delivery target, it lowers to the extent that the package touches the building rooftop. A slight rebound from touching the building rooftop lets the package be released. The energy consumption of a drone is approximately linearly proportional to the payload weight attached to it \cite{dorling2016vehicle}. Therefore, the delivery of packages results in a decrease in the energy consumption of a drone. We repeat this drone navigation process for all the delivery targets. After successfully delivering all the packages, the drone returns to the source location for the next delivery operation.

\bibliographystyle{unsrt}
\bibliography{main}

\end{document}